\pdfoutput=1

\documentclass[11pt,a4paper]{article}
\usepackage[hyperref]{acl2019}
\usepackage{times}
\usepackage{latexsym}
\usepackage{CJKutf8}
\usepackage{graphicx}
\usepackage{xcolor}
\usepackage{amsmath}
\usepackage{amsfonts}
\usepackage{booktabs}
\usepackage{adjustbox}
\usepackage{bm}
\usepackage{float}

\usepackage{pgfplots}
\pgfplotsset{compat=1.14} 

\definecolor{amaranth}{rgb}{0.9, 0.17, 0.31}
\definecolor{kellygreen}{rgb}{77, 186, 23}
\definecolor{azure}{rgb}{0.0, 0.5, 1.0}
\definecolor{gred}{rgb}{0.9, 0.17, 0.31}
\definecolor{gblue}{rgb}{0.0, 0.5, 1.0}
\definecolor{gyellow}{RGB}{244,180,0}
\definecolor{ggreen}{rgb}{0.3, 0.73, 0.09}
\definecolor{ggrey}{RGB}{115,115,115}

\newcommand{\error}[1]{\textcolor{gred}{\textbf{#1}}} 
\newcommand{\fph}[1]{\textcolor{gblue}{\textbf{#1}}} 
\newcommand{\reph}[1]{\textcolor{ggreen}{\textbf{#1}}} 

\aclfinalcopy 

\title{Improving Multi-turn Dialogue Modelling with Utterance ReWriter}

\author{Hui Su$^{1}$\thanks{\hspace{1.5 mm}Both authors contributed equally.}, 
Xiaoyu Shen$^2\footnotemark[1]
$, Rongzhi Zhang$^3$, Fei Sun$^4$, Pengwei Hu$^{5}$\\\textbf{Cheng Niu$^{1}$ and Jie Zhou$^{1}$}\\
$^1$Pattern Recognition Center, Wechat AI, Tencent Inc, China\\
$^2$MPI Informatics \& Spoken Language Systems (LSV), Saarland Informatics Campus\\
$^3$Institute of Software, University of Chinese Academy of Science\\
$^4$Alibaba Group  \hspace{5 mm} $^5$IBM Research, China\\
\tt{aaronsu@tencent.com,xshen@mpi-inf.mpg.de}}

\date{}

\begin{document}
\maketitle
\begin{abstract}
  Recent research has made impressive progress in single-turn dialogue modelling. In the multi-turn setting, however, current models are still far from satisfactory. One major challenge is the frequently occurred coreference and information omission in our daily conversation, making it hard for machines to understand the real intention. In this paper, we propose rewriting the human utterance as a pre-process to help multi-turn dialgoue modelling. Each utterance is first rewritten to recover all coreferred and omitted information. The next processing steps are then performed based on the rewritten utterance. To properly train the utterance rewriter, we collect a new dataset with human annotations and introduce a Transformer-based utterance rewriting architecture using the pointer network. We show the proposed architecture achieves remarkably good performance on the utterance rewriting task. The trained utterance rewriter can be easily integrated into online chatbots and brings general improvement over different domains.\footnote{The code and dataset will be released soon.}
\end{abstract}

\section{Introduction}

Dialogue systems have made dramatic progress in recent years, especially in single-turn chit-chat and FAQ matching~\cite{shang2015neural,ghazvininejad2018knowledge,molino2018cota,chen2019driven}. Nonethless, multi-turn dialogue modelling still remains extremely challenging~\cite{vinyals2015neural,serban2016building,serban2017hierarchical,shen2018nexus,shen2018improving}. The challenge is multi-sided. One most important difficulty is the frequently occurred coreference and information omission in our daily conversations, especially in pro-drop languages like Chinese or Japanese. From our preliminary study of 2,000 Chinese multi-turn conversations, different degrees of coreference and omission exist in more than $70\%$ of the utterances. Capturing the hidden intention beneath them requires deeper understanding of the dialogue context, which is difficult for current neural network-based systems.
\begin{CJK*}{UTF8}{gbsn}
Table \ref{tab:dialog} shows two typical examples in multi-turn dialogues. ``他"(he) from Context 1 is a coreference to ``梅西"(Messi) and ``为什么"(Why) from Context 2 omits the further question of ``为什么最喜欢泰坦尼克"(Why do you like  Tatanic most)?. Without expanding the coreference or omission to recover the full information, the chatbot has no idea how to continue the talk.
\end{CJK*}

    \begin{table}[!t]
    	 \small
    	\centering	
    		\scalebox{0.8}{
    	\begin{tabular}{l|l}
    		\hline
    	   &\textbf{Context 1} \\ 
    	Utterance 1& \emph{Human}:\begin{CJK*}{UTF8}{gbsn} \reph{梅西}有多高？\end{CJK*} \\(Translation)& \emph{Human}: How tall is \reph{Messi}? \\ \hline
    	Utterance 2&	\emph{ChatBot}:\begin{CJK*}{UTF8}{gbsn} 官方说法\reph{他}的身高是5英尺7英寸。\end{CJK*} \\& \emph{ChatBot}: Officially \reph{he} is 5ft 7 inches.\\ \hline
    	Utterance 3&	\emph{Human}:\begin{CJK*}{UTF8}{gbsn} \reph{他}和C罗谁是最好的球员？\end{CJK*} \\& \emph{Human}: Who is the best,  \reph{he} or C.Ronaldo? \\ \hline
    	\error{Utterance $3^\prime$}&  \emph{Human}:\begin{CJK*}{UTF8}{gbsn} \reph{梅西}和C罗谁是最好的球员？\end{CJK*} \\& \emph{Human: Who is the best, \reph{Messi} or C.Ronaldo?} \\ \hline
             &\textbf{Context 2} \\ 
    	Utterance 1& \emph{Human}: \begin{CJK*}{UTF8}{gbsn} 你\fph{最喜欢}什么电影？\end{CJK*} \\& \emph{Human}: What movie do you \fph{like most}? \\ \hline
    	Utterance 2& \emph{ChatBot}:\begin{CJK*}{UTF8}{gbsn} \fph{泰坦尼克}。\end{CJK*} \\& \emph{ChatBot}: \fph{Titanic}.\\ \hline
    	Utterance 3& \emph{Human}:\begin{CJK*}{UTF8}{gbsn} 为什么呢？\end{CJK*} \\& \emph{Human}: Why? \\ \hline
    	\error{Utterance $3^\prime$}& \emph{Human}:\begin{CJK*}{UTF8}{gbsn} 为什么\fph{最喜欢泰坦尼克}？\end{CJK*} \\& \emph{Human}: Why do you \fph{like Titanic most}? \\ \hline
    	\end{tabular}
}
    	\caption{An example of multi-turn dialogue. Each utterance 3 is rewritten into \error{Utterance $3^\prime$}. Green means coreference and blue means omission.}	\vspace{-5mm}
    	\label{tab:dialog}
    \end{table}

To address this concern, we propose simplifying the multi-turn dialogue modelling into a \emph{single-turn} problem by rewriting the current utterance. The utterance rewriter is expected to perform (1) coreference resolution and (2) information completion to recover all coreferred and omitted mentions. In the two examples from Table~\ref{tab:dialog}, each utterance 3 will be rewritten into utterance $3^\prime$. Afterwards, the system will generate a reply by only looking into the utterance $3^\prime$ without considering the previous turns utterance 1 and 2. This simplification shortens the length of dialogue context while still maintaining necessary information needed to provide proper responses, which we believe will help ease the difficulty of multi-turn dialogue modelling. Compared with other methods like memory networks~\cite{sukhbaatar2015end} or explicit belief tracking~\cite{mrkvsic2017neural}, the trained utterance rewriter is model-agnostic and can be easily integrated into other black-box dialogue systems. It is also more memory-efficient because the dialogue history information is reflected in a single rewritten utterance.

To get supervised training data for the utterance rewriting, we construct a Chinese dialogue dataset containing 20k
multi-turn dialogues. Each utterance is paired with corresponding manually annotated rewritings. We model this problem as an extractive generation problem using the Pointer Network~\cite{vinyals2015pointer}. The rewritten utterance is generated by copying words from either the dialogue history or the current utterance based on the attention mechanism~\cite{bahdanau2014neural}. Inspired by the recently proposed Transformer architecture~\cite{vaswani2017attention} in machine translation which can capture better intra-sentence word dependencies, we modify the Transformer architecture to include the pointer network mechanism. The resulting model outperforms the recurrent neural network (RNN) and original Transformer models, achieving an F1 score of over 0.85 for both the coreference resolution and information completion. Furthermore, we integrate our trained utterance rewriter into two online chatbot platforms and find it leads to more accurate intention detection and improves the user engagement.
In summary, our contributions are:
\begin{enumerate}
    \item We collect a high-quality annotated dataset for coreference resolution and information completion in multi-turn dialogues, which might benefit future related research.
    \item We propose a highly effective Transformer-based utterance rewriter outperforming several strong baselines.
    \item The trained utterance rewriter, when integrated into two real-life online chatbots, is shown to bring significant improvement over the original system.
\end{enumerate}
In the next section, we will first go over some related work. Afterwards, in Section 3 and 4, our collected dataset and proposed model are introduced. The experiment results and analysis are presented in Section 5. Finally, some conclusions are drawn in Section 6.
\section{Related Work}
\subsection{Sentence Rewriting}
Sentence rewriting has been widely adopted in various NLP tasks. In machine translation, people have used it to refine the output generations from seq2seq models~\cite{niehues2016pre,junczys2017exploration,grangier2017quickedit,gu2017search}. In text summarization, reediting the retrieved candidates can provide more accurate and abstractive summaries~\cite{see2017get,chen2018fast,cao2018retrieve}. In dialogue modelling, \citet{weston2018retrieve} applied it to rewrite outputs from a retrieval model, but they pay no attention to recovering the hidden information under the coreference and omission. Concurrent with our work, \citet{rastogi2019scaling} adopts a similar idea on English conversations to simplify the downstream SLU task by reformulating the original utterance. Rewriting the source input into some easy-to-process standard format has also gained significant improvements in information retrieval~\cite{riezler2010query}, semantic parsing~\cite{Chen2016SentenceRF} or question answering~\cite{abujabal2018never}, but most of them adopt a simple dictionary or template based rewriting strategy. For multi-turn dialogues, due to the complexity of human languages, designing suitable template-based rewriting rules would be time-consuming.
\subsection{Coreference Resolution}
Coreference resolution aims to link an antecedent for each possible mention. Traditional approaches often adopt a pipeline structure which first identify all pronouns and entities then run clustering algorithms~\cite{haghighi2009simple,lee2011stanford,durrett2013easy,bjorkelund2014learning}. \begin{CJK*}{UTF8}{gbsn}
At both stages, they rely heavily on complicated, fine-grained features. Recently, several neural coreference resolution
systems~\cite{clark2016deep,clark2016improving} utilize distributed representations to reduce human labors. \citet{lee2017end} reported state-of-the-art results with an end-to-end neural coreference resolution system. However, it requires computing the scores for all possible spans, which is computationally inefficient on online dialogue systems. The recently proposed Transformer adopted the self-attention mechanism which could implicitly capture inter-word dependencies in an unsupervised way~\cite{vaswani2017attention}. However, when multiple coreferences occur, it has problems properly distinguishing them. Our proposed architecture is built upon the Transformer architecture, but perform coreference resolution in a supervised setting to help deal with ambiguous mentions.
\section{Dataset}
\label{sec:dataset}
 To get parallel training data for the sentence rewriting, we crawled 200k candidate multi-turn conversational data from several popular Chinese social media platforms for human annotators to work on. Sensitive information is filtered beforehand for later processing. Before starting the annotation, we randomly sample 2,000 conversational data and analyze how often coreference and omission occurs in multi-turn dialogues. Table~\ref{tab:statistics1} lists the statistics. As can be seen, only less than 30\% utterances have neither coreference nor omission and quite a few utterances have both. This further validates the importance of addressing the these situations in multi-turn dialogues. 
 \begin{table}[h]
\centering
\begin{tabular}{lc}
\toprule
& \% Rate \\
\midrule
Coreference&33.5    \\
Omission& 52.4  \\
Neither& 29.7\\
\bottomrule
\end{tabular}
\caption{Proportion of utterances containing coreference and omission in multi-turn conversation}
\label{tab:statistics1}
\end{table}

 In the annotation process, human annotators need to identify these two situations then rewrite the utterance to cover all hidden information. An example is shown in Table~\ref{tab:dialog}. Annotators are required to provide the rewritten utterance $3^\prime$ given the original conversation [utterance 1,2 and 3]. To ensure the annotation quality, 10\% of the annotations from each annotator are daily examined by a project manager and feedbacks are provided. The annotation is considered valid only when the accuracy of examined results surpasses 95\%. Apart from the accuracy examination, the project manage is also required to (1) select topics that are more likely to be talked about in daily conversations, (2) try to cover broader domains and (3) balance the proportion of different coreference and omission patterns. The whole annotation takes 4 months to finish. In the end, we get 40k high-quality parallel samples. Half of them are \emph{negative} samples which do not need any rewriting. The other half are \emph{positive} samples where rewriting is needed. Table~\ref{tab:statistics2} lists the statistics. The rewritten utterance contains 10.5 tokens in average, reducing the context length by 80\%.
 \end{CJK*}

 \begin{table}[h]
\centering
\begin{tabular}{lc}
\toprule
Dataset size:&40,000    \\
Avg. length of original conversation: & 48.8  \\
Avg. length of rewritten utterance: & 10.5  \\
\bottomrule
\end{tabular}
\caption{Statistics of dataset. Length is counted in the unit of Chinese characters.}
\label{tab:statistics2}
\end{table}

\section{Model}
\begin{figure*}[!ht]
\centering
\centerline{\includegraphics[height=8cm]{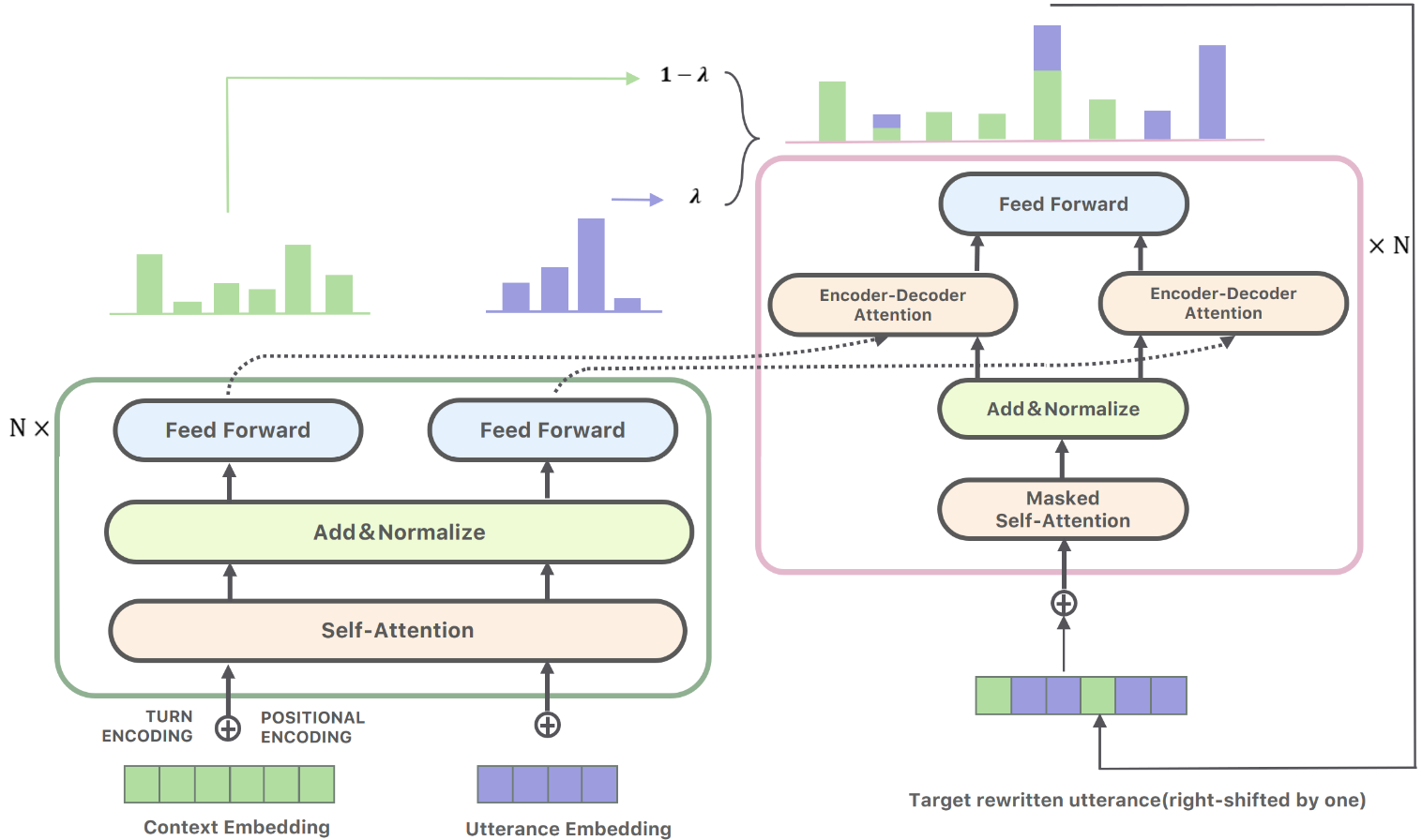}}
\caption{\begin{CJK*}{UTF8}{gbsn}Architecture of our proposed model. Green box is the Transformer encoder and pink box is the decoder. The decoder computes the probability $\lambda$ at each step to decide whether to copy from the context or utterance.\end{CJK*}}
\label{fig:model}
\end{figure*}
\subsection{Problem Formalization}
\begin{CJK*}{UTF8}{gbsn}We denote each training sample as $ (H,U_n \rightarrow R) $. $H = \{U_{1}, U_2,\ldots, U_{n-1}\}$ represents the dialogue history containing the first $n-1$ turn of utterances. $U_{n}$ is the $n$th turn of utterance, the one that needs to be rewritten. $R$ is the rewritten utterance after recovering all corefernced and omitted information in $U_n$. $R$ could be identical to $U_n$ if no coreference or omission is detected (negative sample). Our goal is to learn a mapping function $p(R|(H,U_n))$ that can automatically rewrite $U_n$ based on the history information $H$. The process is to first encode $(H,U_n)$ into s sequence of vectors, then decode $R$ using the pointer network. The next section will explain the steps in order.

\subsection{Encoder}
We unfold all tokens in $(H,U_n)$ into $(w_1, w_2, \ldots, w_{m})$. $m$ is the number of tokens in the whole dialogue. An end-of-turn delimiter is inserted between each two turns. The unfolded sequence of tokens are then encoded with Transformer. We concatenate all tokens in $(H,U_n)$ as the input, in hope that the Transformer can learn rudimentary coreference information within them by means of the self-attention mechanism. For each token $w_i$, the input embedding is the sum of its word embedding, position embedding and turn embedding:
\begin{equation*}
\begin{split}
    &I(w_i) = WE(w_i)+PE(w_i)+TE(w_i)
\end{split}
\end{equation*}
The word embedding $WE(w_i)$ and position embedding $PE(w_i)$ are the same as in normal Transformer architectures~\cite{vaswani2017attention}. We add an additional turn embedding $TE(w_i)$ to indicate which turn each token belongs to. Tokens from the same turn will share the same turn embedding. The input embeddings are then forwarded into $L$ stacked encoders to get the final encoding representations. Each encoder contains a self-attention layer followed by a feedforward neural network.:
\begin{gather*}
\mathbf{E}^{(0)} = \Big[I(w_1),I(w_2),\ldots,I(w_m)\Big]\\
\mathbf{E}^{(l)} = \mathrm{FNN}(\mathrm{MultiHead}(\mathbf{E}^{(l-1)},\mathbf{E}^{(l-1)},\mathbf{E}^{(l-1)}))
\end{gather*}
$\mathrm{FNN}$ is the feedforward neural network and $\mathrm{MultiHead(Q, K, V)}$ is a multi-head attention function taking a query matrix $\mathrm{Q}$, a
key matrix $\mathrm{K}$, and a value matrix $\mathrm{V}$ as inputs. Each self-attention and feedforward component comes with a residual connection and layer-normalization step, which we refer to \citet{vaswani2017attention} for more details. The final encodings are the output from the $L$th encoder $\mathbf{E}^{(L)}$. 
\subsection{Decoder}
The decoder also contains $L$ layers, each layer is composed of three sub-layers. The first sub-layer is
a multi-head self-attention:
\begin{gather*}
\mathbf{M}^l = \mathrm{MultiHead}(\mathbf{D}^{(l-1)},\mathbf{D}^{(l-1)},\mathbf{D}^{(l-1)})
\end{gather*}
$\mathbf{D}^{(0)} = R$. The second sub-layer is encoder-decoder attention that integrates $\mathbf{E}^{(L)}$ into the decoder. In our task, as $H$ and $U_n$ serve different purposes, 
we use separate key-value matrix for tokens coming from the dialogue history $H$ and those coming from $U_n$. The encoded sequence $\mathbf{E}^{(L)}$ obtained from the last section is split into $\mathbf{E}_{H}^{(L)}$ (encodings of tokens from $H$) and $\mathbf{E}_{U_n}^{(L)}$ (encodings of tokens from $U_n$) then processed separately. The encoder-decoder vectors are computed as follows:
\begin{gather*}
\mathbf{C}(H)^l = \mathrm{MultiHead}(\mathbf{M}^{(l)},\mathbf{E}_{H}^{(L)},\mathbf{E}_{H}^{(L)})\\
\mathbf{C}(U_n)^l = \mathrm{MultiHead}(\mathbf{M}^{(l)},\mathbf{E}_{U_n}^{(L)},\mathbf{E}_{U_n}^{(L)})
\end{gather*}
The third sub-layer is a position-wise fully connected feed-forward neural network:
\begin{gather*}
\mathbf{D}^{(l)} = \mathrm{FNN}([\mathbf{C}(H)^l\circ \mathbf{C}(U_n)^l])
\end{gather*}
where $\circ$ denotes vector concatenation.

\subsection{Output Distribution}
In the decoding process, we hope our model could learn whether to copy words from $H$ or $U_n$ at different steps. Therefore, we impose a soft gating weight $\lambda$ to make the decision. The decoding probability is computed by combining the attention distribution from the last decoding layer: 
\begin{gather*}
\begin{split}
    p(R_t {=} w|H, U_n, R_{{<}t}) {=} &\lambda\!\! \sum_{i:(w_i=w)\land (w_i\in \mathrm{H})}\!\!\! a_{t,i}\\ + (1{-}\lambda)\!\!&\sum_{j:(w_j=w)\land (w_j\in U_n)}\!\!\! a^{\prime}_{t,j}
\end{split}\\
a=\mathrm{Attention}(\mathbf{M}^{(L)},\mathbf{E}_{U_n}^{(L)})\\
a^\prime=\mathrm{Attention}(\mathbf{M}^{(L)},\mathbf{E}_{H}^{(L)})\\
\lambda = \sigma \bigl(\bm{w}_{d}^{\top}\mathbf{D}_{t}^{L} + \bm{w}_{H}^{\top}\mathbf{C}(H)^{L}_t + \bm{w}_{U}^{\top}\mathbf{C}(U_n)^{L}_t \bigr)
\end{gather*}
$a$ and $a^\prime$ are the attention distribution over tokens in $H$ and $U_n$ respectively. $\bm{w}_d, \bm{w}_{H}$, and $\bm{w}_{U}$ are parameters to be learned, $\sigma$ is the sigmoid function to output a value between 0 and 1. The gating weight $\lambda$ works like a sentinel to inform the decoder whether to extract information from the dialogue history $H$ or directly copy from $U_n$. If $U_n$ contains neither coreference nor information omission. $\lambda$ would be always 1 to copy the original $U_n$ as the output. Otherwise $\lambda$ becomes 0 when a coreference or omission is detected. The attention mechanism is then responsible of finding the proper coreferred or omitted information from the dialogue history. The whole model is trained end-to-end by maximizing $p(R|H,U_n)$. 



\section{Experiments}
We train our model to perform the utterance rewriting task on our collected dataset. In this section, we focus on answering the following two questions: (1) How accurately our proposed model can perform coreference resolution and information completion respectively and (2) How good the trained utterance rewriter is at helping off-the-shelf dialogue systems provide more appropriate responses. To answer the first question, we compare our models with several strong baselines and test them by both automatic evaluation and human judgement. For the second question, we integrate our rewriting model to two online dialogue systems and analyze how it affects the human-computer interactions. The following section will first introduce the compared models and basic settings, then report our evaluation results.
\subsection{Compared Models}
\label{sec:compared}
When choosing compared models, we are mainly curious to see (1) whether the self-attention based Transformer architecture is superior to other networks like LSTMs, (2) whether the pointer-based generator is better than pure generation-based models and (3) whether it is preferred to split the attention by a coefficient $\lambda$ as in our model. With these intentions, we implement the following four types of models for comparison:
\begin{enumerate}
    \item \textbf{(L/T)-Gen}: Pure generation-based model. Words are generated from a fixed vocabulary.
    \item \textbf{(L/T)-Ptr-Net}: Pure pointer-based model as in \citet{vinyals2015pointer}. Words can only be copied from the input.
    \item \textbf{(L/T)-Ptr-Gen}: Hybrid pointer+generation model as in \citet{see2017get}. Words can be either copied from the input or generated from a fixed vocabulary.
    \item \textbf{(L/T)-Ptr-$\lambda$}: Our proposed model which split the attention by a coefficient $\lambda$.
\end{enumerate}
(L/T) denotes the encoder-decoder structure is the LSTM or Transformer. For the first three types of models, we unfold all tokens from the dialogue as the input. No difference is made between the dialogue history and the utterance to be rewritten.

\subsection{Experiment Settings}
\begin{table*}
  \centering
  \begin{adjustbox}{max width=\textwidth}
  \begin{tabular}{l c c c c c c c } \toprule
    &  BLEU-1 & BLEU-2 & BLEU-4 & ROUGE-1 & ROUGE-2 & ROUGE-L & EM \\ \midrule
    L-Gen & 65.49 & 55.38 & 38.69 & 65.57 & 48.57 & 66.38 & 47.14$\vert$80.18 \\
    L-Ptr-Gen & 69.78 & 59.25  & 43.07 & 68.24 & 54.13 & 70.36 & 47.35$\vert$84.09 \\
    L-Ptr-Net & 71.70 & 60.29 & 44.72 & 70.81 & 56.35 & 72.33 & 48.24$\vert$91.94 \\
    L-Ptr-$\lambda$ & 72.26  & 62.15 & 47.11 & 73.47 & 57.51 & 74.55 & 51.66$\vert$93.01 \\ \midrule
    T-Gen & 68.74 & 59.09 & 42.57 & 69.12 & 50.92 & 69.70 & 48.59$\vert$87.61 \\
    T-Ptr-Gen & 70.67 & 62.80 & 45.17 & 73.96 & 53.14 & 72.07 & 49.86$\vert$89.62 \\
    T-Ptr-Net & 75.10 & 66.89 & 48.11 & 76.10 & 58.51 & 75.54 & 53.30$\vert$94.71 \\
    T-Ptr-$\lambda$ & \textbf{77.85}  & \textbf{68.21} & \textbf{52.47} & \textbf{78.49}  & \textbf{60.53} & \textbf{77.70}  & \textbf{55.84}$\vert$\textbf{98.14} \\
    \bottomrule
  \end{tabular}
  \end{adjustbox}
    \caption{BLEU, ROUGE (F$_1$), and EM scores on the test set. EM score is split into the results on the positive (left) and negative (right) test samples. The first half is LSTM-based models and the second half is Transformer-based. 
  Bold denotes best results.}
   \label{tab:full}
\end{table*}

\paragraph{Transformer-based models}
We set the hidden size as 512. 
The attention has 8 individual heads and the encoder/decoder have 6 individual stacked layers. Models are optimized with the Adam optimizer. The initial learning rate is 0.0001 and batch size is 64. All hyperparameters are tuned base on the performance on the validation data.

\paragraph{LSTM-based Models} We encode words with a single-layer bidirectional LSTM and decode with a uni-directional LSTM. We use 128-dimensional word embeddings and 256-dimensional hidden states for both the encoder and decoder.\footnote{We tried increasing the dimension but find it degrades the performance.} The
batch size is set as 128. Models are trained using Adagrad with learning
rate 0.15 and initial accumulator value 0.1, same as in ~\citet{see2017get}.

\paragraph{General Setup} We built our vocabulary based on character-based segmentation for Chinese scripts. For non-Chinese characters, like frequently mentioned entity names ``Kobe" and ``NBA", we split them by space and keep all unique tokens which appear more than twice. The resulting vocabulary size is
5629 (4813 Chinese characters and 816 other tokens), including the end-of-turn delimiter and a special UNK token for all unknown words. In the testing stage, all models decode words by beam search with beam size set to 4.

\subsection{Quality of Sentence ReWriting}
\begin{table}[h]
\centering
\begin{tabular}{lccc}
\toprule
 & Precision & Recall & F1\\
\midrule
\citet{lee2017end} & 0.82 & 0.78 & 0.80\\\midrule
L-Gen  & 0.76 & 0.66 & 0.71\\
L-Ptr-Gen  & 0.81 & 0.76 & 0.78   \\
L-Ptr-Net & 0.83 & 0.78 & 0.81 \\
L-Ptr-$\lambda$   & 0.85  & 0.82 & 0.83\\ \midrule
T-Gen  & 0.80 & 0.75 & 0.77\\
T-Ptr-Gen  & 0.85 &0.81  & 0.83\\
T-Ptr-Net  & 0.88 & 0.87 & 0.88\\
T-Ptr-$\lambda$  & \textbf{0.93} & \textbf{0.90} & \textbf{0.92}\\
\bottomrule
\end{tabular}
\caption{Precision, recall and F1 score of coreference resolution. First row is the current state-of-the-art coreference resolution model}
\label{tab:coref}
\end{table}
\paragraph{Accuracy of Generation}
We first evaluate the accuracy of generation
leveraging three metrics: BLEU, ROUGE, and the exact match score(EM) (the percentage of decoded sequences that exactly match the human references). For the EM score, we report separately on the positive and negative samples to see the difference. We report BLEU-1, 2, 4 scores and the F1 scores of ROUGE-1, 2, L.
The results are listed in Table~\ref{tab:full}. We can have several observations in response to the three questions proposed in the beginning of Section~\ref{sec:compared}:
 \begin{enumerate}
     \item Transformer-based models lead to significant improvement compare with LSTM-based counterparts. This implies the self-attention mechanism is helpful in identifying coreferred and omitted information. More analysis on how it helps coreference resolution can be seen in the next section.
     \item  The generation mode does not work well in our setting since all words can be retrieved from either $H$ or $U_n$. Pointer-based models outperform the more complex generation-based and hybrid ones.
     \item Separately processing $H$ and $U_n$ then combine their attention with a learned $\lambda$ performs better than treating the whole dialogue tokens as s single input, though the improvement is less significant compared with previous two mentions.
 \end{enumerate}
Overall our proposed model achieves remarkably good performance, with $55.84\%$ of its generations exactly matches the human reference on the positive samples. For negative samples, our model properly copied the the original utterances in $98.14\%$ of the cases. It suggests our model is already able to identify the utterances that do not need rewriting. Future work should work on improving the rewriting ability on positive samples.

\begin{figure}
\centering
\centerline{\includegraphics[width=7.5cm]{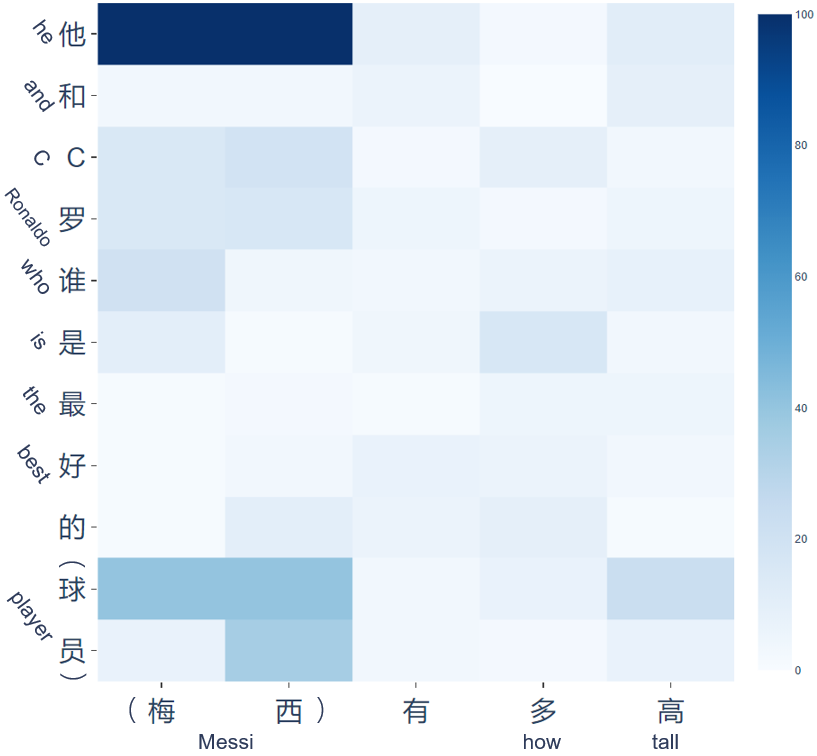}}
\caption{Visualization of the self-attention weights in Transformer. ``他''(he) is properly aligned to ``梅西''(Messi).}
\label{fig:core}
\end{figure}

\begin{table*}
	\centering
	
	\begin{adjustbox}{max width=\textwidth}
	\renewcommand{\arraystretch}{1}
	\begin{tabular}{l l l } \toprule
 History&  \begin{CJK*}{UTF8}{gbsn}U1: 你看莎士比亚吗~U2: 特别喜欢罗密欧与朱丽叶\end{CJK*} & \begin{CJK*}{UTF8}{gbsn}U1: 你玩英雄联盟吗~U2: 是的 \end{CJK*}\\ 
     (Translation) & U1: Do you read Shakespeare ~U2: I especially like Romeo and Juliet & U1: Do you play League of Legends~U2: Yes. \\ \midrule 
     Utterance & \begin{CJK*}{UTF8}{gbsn}U3:喜欢哪个角色\end{CJK*} & \begin{CJK*}{UTF8}{gbsn}U3: 什么时候开始的\end{CJK*} \\ 
      & U3: Which character do you like                               & U3: When did it start \\ \midrule %
     Ground Truth & \begin{CJK*}{UTF8}{gbsn}你喜欢罗密欧与朱丽叶哪个角色\end{CJK*}   &  \begin{CJK*}{UTF8}{gbsn}什么时候开始玩英雄联盟的 \end{CJK*} \\ 
      & Which character do you like in Romeo and Juliet  &  When did you start to play League of Legends \\ \midrule
L-Gen & \begin{CJK*}{UTF8}{gbsn}你喜欢莎士比亚吗\end{CJK*} // Do you like Shakespeare & \begin{CJK*}{UTF8}{gbsn}什么时候开始开始开始\end{CJK*} // When start start start\\
   L-Ptr-Gen & \begin{CJK*}{UTF8}{gbsn}你喜欢罗密欧角色角色\end{CJK*} // You like Romeo character character & \begin{CJK*}{UTF8}{gbsn}什么时候开始的\end{CJK*} // When did it start  \\
   L-Ptr-Net & \begin{CJK*}{UTF8}{gbsn}你喜欢罗密欧与朱丽叶\end{CJK*} // You like Romeo and Juliet &  \begin{CJK*}{UTF8}{gbsn}什么时候英雄联盟开始的\end{CJK*} // When did League of Legends start\\
   L-Ptr-$\lambda$ & \begin{CJK*}{UTF8}{gbsn}你喜欢罗密欧与朱丽叶角色\end{CJK*} // You like Romeo and Juliet character & \begin{CJK*}{UTF8}{gbsn}\error{什么时候开始玩英雄联盟的}\end{CJK*} // \error{When did you start to play League of Legends}\\
   T-Gen & \begin{CJK*}{UTF8}{gbsn}你喜欢罗密欧与朱丽叶\end{CJK*} // You like Romeo and Juliet & \begin{CJK*}{UTF8}{gbsn}是的 什么时候开始玩的\end{CJK*} // Yes When start to play\\ 
   T-Ptr-Gen & \begin{CJK*}{UTF8}{gbsn}你喜欢罗密欧与朱丽叶哪个\end{CJK*} // Which do you like in Romeo and Juliet  & \begin{CJK*}{UTF8}{gbsn}什么时候开始的\end{CJK*} // When did it start\\
   T-Ptr-Net   & \begin{CJK*}{UTF8}{gbsn}你喜欢罗密欧与朱丽叶角色\end{CJK*}  //  Character you like Romeo and Juliet & \begin{CJK*}{UTF8}{gbsn}英雄联盟什么时候开始玩的\end{CJK*} // League of Legends When did you start to play \\
   T-Ptr-$\lambda$  & \begin{CJK*}{UTF8}{gbsn}\error{你喜欢罗密欧与朱丽叶哪个角色}\end{CJK*} // \error{Which character do you like Romeo and Juliet} & \begin{CJK*}{UTF8}{gbsn}\error{什么时候开始玩英雄联盟的}\end{CJK*} // \error{When did you start to play League of Legends}\\
    \bottomrule
  \end{tabular}
  \end{adjustbox}
  \caption{Examples of rewritten utterances. Highlighted utterances are exactly the same as the ground truth.}
    \label{tab:cs}
\end{table*}

\paragraph{Coreference Resolution}
Apart from the standard metrics for text generation, we specifically test the precision, recall and F1 score of coreference resolution on our task. A pronoun or a noun is considered as properly coreferred if the rewritten utterance contains the correct mention in the corresponding referent. The result is shown in Table~\ref{tab:coref}. To compare with current state-of-the-art models. We train the model from \citet{lee2017end} on our task and report the results on the first row. The result is quite consistent with the findings from the last section. Our final model outperforms the others by a large margin, reaching a precision score of 93\% and recall score of 90\%. It implies our model is already quite good at finding the proper coreference. Future challenges would be more about information completion.
\begin{CJK*}{UTF8}{gbsn}
Figure~\ref{fig:core} further provides an examples of how the Transformer can help implicitly learn the coreference resolution through the self-attention mechanism. The same example is also shown in Table~\ref{tab:dialog}. The pronoun ``他''(he) in the utterance is properly aligned to the mention ``梅西''(Messi) in the dialogue history, also partially to ``球员''(player) which is the occupation of him. The implicitly learned coreference relation should be part of the reason that Transformers outperform LSTM models on the coreference resolution task.
\end{CJK*}

\begin{table}[h]
	\centering
	
	\setlength{\tabcolsep}{7pt}
	\begin{adjustbox}{max width=\linewidth}
	\begin{tabular}{l r c c c } \toprule
    Model & Recall  & Precision & F1 & Fluency  \\ \midrule
    L-Gen  & 0.65 & 0.70 &0.67& 4.31\\
    L-Ptr-Gen  & 0.70 & 0.74 &0.72& 4.52   \\
    L-Ptr-Net & 0.78 & 0.81 &0.79& 4.74 \\
    L-Ptr-$\lambda$   & 0.80  & 0.82  & 0.81& 4.82\\ \midrule
    T-Gen  & 0.71 & 0.74 &0.73& 4.74\\
    T-Ptr-Gen  & 0.77 & 0.81  &0.79& 4.85\\
    T-Ptr-Net  & 0.82 & 0.84 &0.83 &4.87\\
    T-Ptr-$\lambda$  & \textbf{0.85} & \textbf{0.87}  & \textbf{0.86}& \textbf{4.90}\\\midrule
   Human & - & - &-& 4.97 \\
    \bottomrule
  \end{tabular}
  \end{adjustbox}
  \caption{Recall, Precision, F1 score on information completion and Human evaluation results on fluency.}
  \label{tab:manual}
\end{table}
\paragraph{Information Completion}
\begin{table*}[ht]
	\centering

	\begin{adjustbox}{max width=\textwidth}
	\renewcommand{\arraystretch}{1}
	\begin{tabular}{l l l } 
	\toprule
	\multicolumn{3}{l}{Task-Oriented Chatbot}\\ \midrule
	Context & \multicolumn{2}{l}{\begin{CJK*}{UTF8}{gbsn}U1: \fph{北京天气}怎么样~~U2: 天气晴朗，温度适宜\end{CJK*}} \\
    (Translation) & \multicolumn{2}{l}{ U1:~How is the \fph{weather in Beijing}~~U2: The weather is fine and the temperature is suitable } \\
    Utterance & \begin{CJK*}{UTF8}{gbsn}U3: 那穿什么衣服合适\end{CJK*} & \begin{CJK*}{UTF8}{gbsn}U3: \fph{北京天气}穿什么合适\end{CJK*} \\
     &U3:~Then what clothes are suitable to wear & U3:~ What clothes are suitable for \fph{weather in Beijing} \\
    Intention &\begin{CJK*}{UTF8}{gbsn}生活购物 \end{CJK*}& \begin{CJK*}{UTF8}{gbsn}城市天气\end{CJK*} \\
    & Life Shopping & City Weather  \\
    Chatbot Answer& \begin{CJK*}{UTF8}{gbsn}您想要购买什么类型的衣服\end{CJK*} & \begin{CJK*}{UTF8}{gbsn}根据天气推荐穿一件外套\end{CJK*} \\
    & What type of clothes do you want to buy  & You'd better wear a coat according to the weather \\\midrule
    \multicolumn{3}{l}{Chit-Chat Chatbot}\\ \midrule
    	Context & \multicolumn{2}{l}{\begin{CJK*}{UTF8}{gbsn}U1: 库里的三分真准啊~~U2: \fph{勇士今年又是冠军}\end{CJK*}} \\
     & \multicolumn{2}{l}{ U1:~Curry's 3-pointer is really good~~U2: \fph{The Warriors are the champion again this year} } \\
    Utterance & \begin{CJK*}{UTF8}{gbsn}U3: 我也觉得\end{CJK*} & \begin{CJK*}{UTF8}{gbsn}U3: 我也觉得\fph{勇士今年又是冠军}\end{CJK*} \\
     &U3:~I agree & U3:~ I agree that \fph{the Warriors are the champion again this year} \\
    Chatbot Answer& \begin{CJK*}{UTF8}{gbsn}觉得什么\end{CJK*} & \begin{CJK*}{UTF8}{gbsn}勇士真的厉害啊\end{CJK*} \\
    & agree what  & The Warriors are so strong \\
    \bottomrule
  \end{tabular}
  \end{adjustbox}
  	\caption{Examples of integrated test. Left column is the original system and right is the one with utterance rewriter. Blue words denote completed information by the utterance rewriter.}
    \label{tab:d}
\end{table*}

\begin{table}[h]
\centering
\begin{tabular}{lcc}
\toprule
 Model & Intention Precision & CPS\\
\midrule
Original & 80.77 &6.3\\
With Rewrite  & 89.91&7.7 \\
\bottomrule
\end{tabular}
\caption{Results of integrated testing. Intention precision for task-oriented and conversation-turns-per-session (CPS) for chitchat.}
\label{tab:int}
\end{table}
Similar as coreference resolution, we evaluate the quality of information completeness separately. One omitted information is considered as properly completed if the rewritten utterance recovers the omitted words. Since it inserts new words to the original utterance, we further conduct a human evaluation to measure the fluency of rewritten utterances. We randomly sample 600 samples from our positive test set. Three participants were asked to judge whether the rewritten utterance is a fluent sentence with the score 1(not fluent)-5(fluent). The fluency score for each model is averaged over all human evaluated scores. 

The results are shown in Table~\ref{tab:manual}. Basically the condition is similar as in Table~\ref{tab:coref}. T-Ptr-$\lambda$ achieves the best performance, with the F1 score of 0.86. The performance is slightly worse than coreference resolution since information omission is more implicit. Retrieving all hidden information is sometimes difficult even for humans. Moreover, the fluency of our model's generations is very good, only slightly worse than the human reference (4.90 vs 4.97). Information completeness does not have much effects on the fluency. Examples of rewritten utterances are shown in Table~\ref{tab:cs}.

\subsection{Integration Testing}
In this section, we study how the proposed utterance rewriter can be integrated into off-the-shelf online chatbots to improve the quality of generated responses. We use our best model T-Ptr-$\lambda$ to rewrite each utterance based on the dialogue context. The rewritten utterance is then forwarded to the system for response generation. We apply on both a task-oriented and chitchat setting. The results are compared with the original system having no utterance rewriter.

\paragraph{Task-oriented}
Our task-oriented dialogue system contains an intention classifier built on FastText\cite{bojanowski2017enriching} and a set of templates that perform policy decision and slot-value filling sequentially. Intention detection is a most important component in task-oriented dialogues and its accuracy will affect all the following steps. We define 30 intention classes like weather, hotel booking and shopping. The training data contains 35,447 human annotations. 
With the combination of our rewriter, the intention classier is able to achieve a precision of 89.91\%, outperforming the original system by over 9\%. The improved intention classification further lead to better conversations. An example is shown in Table~\ref{tab:d}, a multi-turn conversation about the weather. The user first asks ``How is the weather in Beijing", then follows with a further question about ``Then what clothes are suitable to wear". The original system wrongly classified the user intention as shopping since this is a common conversational pattern in shopping. In contrast, our utterance rewriter is able to recover the omitted information ``under the weather in Beijing". Based on the rewritten utterance, the classifier is able to correctly detect the intention and provide proper responses.

\paragraph{Chitchat}
Our social chatbot contains two separate engines for multi-turn and single-turn dialogues. Each engine is a hybrid retrieval and generation model. In real-life applications, a user query would be simultaneously distributed to these two engines. The returned candidate responses are then reranked to provide the final response. Generally the model is already able to provide rather high-quality responses under the single-turn condition, but under multi-turn conversations, the complex context dependency makes the generation difficult. We integrate our utterance rewriter into the single-turn engine and compare with the original model by conducting the online A/B test. Specifically, we randomly split the users into two groups. One talks with the original system and the other talks with the system integrated with the utterance rewriter. All users are unconscious of the details about our system. The whole test lasted one month. Table~\ref{tab:int} shows the Conversation-turns Per Session (CPS), which is the average number of conversation-turns between the chatbot and the user in a session. The utterance rewriter increases the average CPS from 6.3 to 7.7, indicating the user is more engaged with the integrated model. Table~\ref{tab:d} shows an example of how the utterance rewriter helps with the generation. After the rewriting, the model can better understand the dialogue is about the NBA team Warriors, but the original model feels confused and only provides a generic response.

 \end{CJK*}
 
 \section{Conclusion}
In this paper, we propose improving multi-turn dialogue modelling by imposing a separate utterance rewriter. The rewriter is trained to recover the coreferred and omitted information of user utterances. We collect a high-quality manually annotated dataset and designed a Transformer-pointer based architecture to train the utterance rewriter. The trained utterance rewriter performs remarkably well and, when integrated into two online chatbot applications, significantly improves the intention detection and user engagement. We hope the collected dataset and proposed model can benefit future related research.

\section*{Acknowledgments}
We thank all anonymous reviewers and the dialogue system team of Wechat AI for valuable comments. Xiaoyu Shen is supported by IMPRS-CS
fellowship.
\bibliography{acl2019}
\bibliographystyle{acl_natbib}

\end{document}